\documentclass{article} 
\usepackage[numbers]{natbib}
\usepackage[final]{nips_2017}
\pdfoutput=1
\usepackage[dvipsnames]{xcolor}
\usepackage[utf8]{inputenc} 
\usepackage[T1]{fontenc}    
\usepackage{amsmath}        
\usepackage{hyperref}       
\usepackage{url}            
\usepackage{booktabs}       
\usepackage{amsfonts}       
\usepackage{nicefrac}       
\usepackage{microtype}      
\usepackage{graphicx}
\usepackage{wrapfig}
\usepackage{paralist}

\usepackage{algorithm}
\usepackage[noend]{algorithmic}
\renewcommand{\algorithmiccomment}[1]{\bgroup\hfill $\triangleright$ ~#1\egroup}

\definecolor{ColorCma}{rgb}{0.8,1,1}
\definecolor{ColorMa}{rgb}{1,0.7,1}
\definecolor{ColorLmma}{rgb}{0.35,0.7,0.35}
\newcommand{\cmacolor}{ColorCma}
\newcommand{\macolor}{ColorMa}
\newcommand{\lmmacolor}{ColorLmma}

\newcommand{\cma}[1]{\colorbox{\cmacolor}{$#1$}}
\newcommand{\cmatext}[1]{\colorbox{\cmacolor}{#1}}
\newcommand{\maes}[1]{\colorbox{\macolor}{$#1$}}
\newcommand{\maestext}[1]{\colorbox{\macolor}{#1}}
\newcommand{\lmma}[1]{\colorbox{\lmmacolor}{$#1$}}
\newcommand{\lmmatext}[1]{\colorbox{\lmmacolor}{#1}}

\usepackage[font=small,labelfont=bf]{caption}

\title{Limited-Memory Matrix Adaptation\\for Large Scale Black-box Optimization}

\def\R{{\rm I\hspace{-0.50ex}R}}

\newcommand{\vc}[1]{\textit{\textbf{#1}}}

\newcommand{\ma}[1]{\mathchoice{\mbox{\boldmath$\displaystyle#1$}}
  {\mbox{\boldmath$\textstyle#1$}} {\mbox{\boldmath$\scriptstyle#1$}}
  {\mbox{\boldmath$\scriptscriptstyle#1$}}}
\renewcommand{\ma}[1]{\mathnormal{\mathbf{#1}}}
\newcommand{\mstr}[1]{\mathrm{#1}}
\newcommand{\C}{ \ensuremath{\ma{C}} }
\newcommand{\I}{ \ensuremath{\ma{I}} }
\newcommand{\M}{ \ensuremath{\ma{M}} }

\newcommand{\dd}{n}

\def\Id{\ensuremath{\ma{I}}}

\author{
	Ilya Loshchilov\\
	Research Group on Machine Learning\\for Automated Algorithm Design\\
	University of Freiburg, Germany\\
	\texttt{ilya.loshchilov@gmail.com}
	\And
	Tobias Glasmachers\\
	Institut f\"ur Neuroinformatik\\
	Ruhr-Universit\"at Bochum, Germany\\
	\texttt{tobias.glasmachers@ini.rub.de}
	\And
	Hans-Georg Beyer\\
	Research Center Process and Product Engineering\\
	Vorarlberg University of Applied Sciences, Dornbirn, Austria\\
	\texttt{hans-georg.beyer@fhv.at}
}

\begin{document}

\maketitle

\begin{abstract}
The Covariance Matrix Adaptation Evolution Strategy (CMA-ES) is a popular method to deal with nonconvex and/or stochastic optimization problems when the gradient information is not available. 
Being based on the CMA-ES, 
the recently proposed Matrix Adaptation Evolution Strategy (MA-ES) provides a rather surprising result that the covariance matrix and all associated operations (e.g., potentially unstable eigendecomposition) can be replaced in the CMA-ES by a updated transformation matrix without any loss of performance. In order to further simplify MA-ES and reduce its  $\mathcal{O}\big(n^2\big)$ time and storage complexity to $\mathcal{O}\big(n\log(n)\big)$, we  present the Limited-Memory Matrix Adaptation Evolution Strategy (LM-MA-ES) for efficient zeroth order large-scale optimization. The algorithm demonstrates state-of-the-art performance on a set of established large-scale benchmarks. We explore the algorithm on the problem of generating adversarial inputs for a (non-smooth) random forest classifier, demonstrating a surprising vulnerability of the classifier.




\end{abstract}

\section{Introduction}

Evolution Strategies (ESs) are optimization methods originally inspired by mutation of organic beings and designed to establish ``a reward-based system, to increase the probability of those changes, which lead to improvements of quality of the system'' \citep{Rastrigin1963}. 
Going far beyond their biologically inspired roots, they have been developed into state-of-the-art zeroth order search methods \citep{hansen2015evolution}.
Evolution Strategies \citep{1973RechenbergEvolutionsstrategie} consider an objective function $f: \R^{\dd} \mapsto \R, \vc{x} \mapsto f(\vc{x})$ to be minimized by sampling $i \in \{1,\dots,\lambda\}$ candidate solutions at iteration $t$ as
\begin{equation}
\label{sampling1}
  \vc{x}^{(t)}_i \leftarrow \vc{y}^{(t)} + \sigma^{(t)} \cdot {\mathcal N}  \hspace{-0.13em}\left({\ma{0},{\C}^{(t)}}\right),
\end{equation}
where $\vc{y}^{(t)}$ is the current estimate of the optimum, $\C^{(t)} \in \R^{n \times n}$ is a covariance matrix initialized to the identity matrix $\I$, and $\sigma^{(t)}$ is a scaling factor for the mutation step, often referred to as the global step size.
Both $\vc{y}^{(t)}$ and $\sigma^{(t)}$ are to be adapted or \emph{learned} over time in Evolution Strategies. Modern ESs such as the Covariance Matrix Adaptation Evolution Strategy (CMA-ES) also include the adaptation of $\C^{(t)}$ \citep{hansen1996adapting,2003HansenCMA} to the shape of the local landscape, resembling second order methods. Recent theoretical studies of ES and CMA-ES from the prospective of information geometry \citep{wierstra2014natural, akimoto2010bidirectional, 2017OllivierIGO, beyer2014convergence}, connecting the method to natural gradient learning, have made significant progress in understanding the principles underpinning the state-of-the-art performance of the algorithm~\citep{2009HansenBBOBcomparing31algo}.
The variety of algorithms  \citep{hansen2015evolution} derived from and inspired by the theoretical studies helped to notice that the core component of CMA-ES, the covariance matrix itself (and covariance matrix square root operations) can be removed from the algorithm without any loss of performance~\citep{beyer2017simplify}.%
\footnote{Of course, the transformation matrix needed for sampling the multivariate Gaussian is kept, enabling variable metric optimization.}
The final algorithm called Matrix Adaptation Evolution Strategy (MA-ES ~\citep{beyer2017simplify}) is conceptually simpler and involves only matrix-matrix and matrix-vector operations, which, however, lead to $\mathcal{O}\big(n^3/\log(n)\big)$ time complexity per sample (here, we show its $\mathcal{O}\big(n^2\big)$ implementation), and $\mathcal{O}(n^2)$ space complexity.

In machine learning, ESs are mainly used for direct policy search in reinforcement learning \citep{gomez2008accelerated, heidrich2009hoeffding, stulp2013robot, salimans2017evolution}, hyperparameter tuning in supervised learning, 
e.g., for Support Vector Machines \citep{glasmachers2008uncertainty,igel2011evolutionary} and Deep Neural Networks \citep{loshchilov2016cma}. With the steadily increasing dimensionality of real-world optimization problems, the new challenges of large-scale black-box optimization become more pronounced for CMA-ES and MA-ES due to their $\mathcal{O}(n^2)$ complexity. 
To address them, a number of large-scale CMA-ES variants has been proposed   \citep{knight2007reducing,ros2008simple,sun2011linear,akimoto2014comparison,loshchilov2017lm,akimoto2016online} including the Limited-Memory CMA-ES (LM-CMA-ES  \citep{loshchilov2017lm}) that matches the performance of quasi-Newton methods such as L-BFGS \citep{1970ShannoBFGS} when dealing with large-scale black-box problems at a moderate cost of $O(n\log(n))$ time and space complexity.

In this work, we combine the best of two worlds: inspired by LM-CMA-ES we present the Limited-Memory Matrix Adaptation Evolution Strategy (LM-MA-ES), which matches state-of-the-art results while reducing the time and space complexity of MA-ES to $O(n\log(n))$ per sample.

\section{Variable Metric Evolution Strategies: CMA-ES and MA-ES}

Our discussion on variable metric ESs is based on Algorithm \ref{ALLALGOS}, which highlights the similarities and differences between CMA-ES, MA-ES, and the proposed LM-MA-ES.

The sampling of the $\lambda$ candidate solutions in CMA-ES is described by eq.~\eqref{sampling1} and involves a matrix-vector product between a matrix $\sqrt{\C}$ and a vector $\vc{z}_i$ sampled from the $n$-dimensional standard normal distribution. This operation (see line \ref{CMAdi} in Algorithm \ref{ALLALGOS}) requires $\sqrt{\C}$ to be stored (hence, the quadratic in $n$ space cost) and the matrix-vector multiplication to be performed (hence, the quadratic in $n$ time cost). The resulting vector $\vc{d}_i$ represents a direction of the so-called mutation operation. The $i$-th candidate solution is obtained by changing (mutating) the current estimate of the optimum $\vc{y}$ by $\vc{d}_i$ multiplied by the global mutation step-size $\sigma$ (line~\ref{GenerateEnd}). The rationale behind parameterizing the sampling distribution by ${\mathcal N}  \hspace{-0.13em}\left({\vc{y}},\sigma^2{\C}\right)$ and not just ${\mathcal N}  \hspace{-0.13em}\left({\vc{y}},{\C}\right)$, i.e., decoupling ${\C}$ and $\sigma$, lies in the observation that 
 $\sigma$ can be learned more quickly and more robustly than ${\C}$ and its adaptation alone enables linear convergence on scale-invariant problems \citep{jaegerskupper2005rigorous}. 

ESs are invariant to rank-preserving / strictly monotonic transformations of $f$-values because all operations are based on ranks of evaluated solutions. The estimate of the optimum $\vc{y}$ is updated by a weighted sum of mutation steps taken by the top ranked $\mu$ out $\lambda$ solutions (line~\ref{ComputeNewMean}). The recent analysis  \citep{akimoto2017quality} demonstrated that the optimal recombination weights $\vc{w}$ for Sphere function (see Table~1) are independent of the Hessian matrix, and hence optimal for all convex quadratic functions.

The currently most commonly applied adaptation rule for the step size is the cumulative step-size adaptation (CSA) mechanism~\citep{2001HansenCMA}. It is based on the length of an evolution path $\vc{p}_{\sigma}$, an exponentially fading record of recent most successful steps $\vc{z}_{i:\lambda}$ (see line \ref{SigmaPathUpdate}). If the path becomes too long (the expected path length of a Gaussian random walk can be approximated by $\sqrt{n}$ when $n$ is large), indicating that recent steps tend to move into the same direction, then the step size is increased. On the contrary, a too short path indicating oscillations due to overjumping the optimum results in a reduction of the step size. Rigorous analysis of CSA with and without cumulation is given in \citep{BH14}.

The seminal CMA-ES algorithm \citep{hansen1996adapting,2003HansenCMA} introduced adaptation of the covariance matrix, which renders the algorithm invariant to linear transformations of the search space (achieved in practice after an initial adaptation phase) and hence enables a fast convergence rate independent of the conditioning of the problem, resembling second order methods. The covariance matrix is adapted towards a weighted maximum likelihood estimate of the $\mu$ most successful samples (rank-$\mu$ update) with learning rate $c_{\mu}$ and a second evolution path (rank-$1$ update) with learning rate $c_1$ (see line \ref{cmaCupdate}); this update has an alternative interpretation as a stochastic gradient step on the information geometric manifold forming the algorithm's state space  \citep{akimoto2010bidirectional,2017OllivierIGO}. While the default hyperparameter values of CMA-ES given in Algorithm \ref{ALLALGOS} are known to be robust, their optimal values can be adapted online during the optimization process attempting to provide an additional level of invariance \citep{loshchilov2014maximum}. 

Most  implementations of CMA-ES consider eigendecomposition procedures of $\mathcal{O}(n^3)$ time complexity per call to obtain $\sqrt{\C}$ from $\C$ only every $n/\lambda$ iterations (see line \ref{CMAdi}) to achieve amortized $\mathcal{O}(n^2)$ time complexity per sampled solution. Numerical stability of the $\mathcal{O}(n^2)$ update can be ensured by maintaining a triangular Cholesky factor~\citep{krause2016cma}.

The recently proposed Matrix Adaptation Evolution Strategy (MA-ES) greatly simplifies CMA-ES by avoiding the construction of the covariance matrix $\C$. Instead it maintains only a transformation matrix $\M$ representing $\sqrt{\C}$. After removing the approximate redundancy of $\vc{p}_{\sigma}$ and $\vc{p}_{c}$, $\M$ can be updated multiplicatively (line~\ref{MAupdate}). Matrix multiplication is an $\mathcal{O}(n^3)$ operation, therefore we propose to replace the multiplicative update at iteration $t$ by the equivalent additive update
\begin{equation}
\label{MAsampling12}
\M^{(t+1)} \leftarrow \left(1 - \frac{c_{1}}{2} - \frac{c_{\mu}}{2}\right) \M^{(t)} + \frac{c_{1}}{2} \vc{d}^{(t)}_{\sigma}(\vc{p}^{(t)}_{\sigma})^T + \frac{c_{\mu}}{2} \sum_{i=1}^{\mu} w_i \vc{d}^{(t)}_{i:\lambda}(\vc{z}^{(t)}_{i:\lambda})^T,
\end{equation}
which achieves $\mathcal{O}(n^2)$ time cost thanks to precomputing $\vc{d}^{(t)}_{\sigma}=\M^{(t)} \vc{p}^{(t)}_{\sigma}$ and reusing $\vc{d}^{(t)}_{i:\lambda}$ vectors. The resulting algorithm is referred to as \emph{fast MA-ES}. 

\begin{algorithm}[tb!]
\caption{\cmatext{CMA-ES}, \maestext{MA-ES} and \lmmatext{LM-MA-ES}}
\label{ALLALGOS}
\begin{algorithmic}[1]
\STATE{\textbf{given} $n \in \mathbb{N}_+$, $\lambda = 4 + \lfloor 3 \ln \, n  \rfloor $, $\mu =  \lfloor \lambda/2   \rfloor $, 
											$w_i = \frac{ \ln(\mu + \frac{1}{2}) - \ln\,i}{ \sum^{\mu}_{j=1}(\ln(\mu + \frac{1}{2})-\ln\,j)} \; \mstr{for} \; i=1, \ldots, \mu$,
											$\mu_w = \frac{1}{\sum^{\mu}_{i=1} w^2_i}$, 
						$\cma{c_{\sigma} = \frac{\mu_w + 2}{n+\mu_w+5}, c_{c} = \frac{4}{n + 4}, c_1 = \frac{2}{(n+1.3)^2 +\mu_w}, c_{\mu} = \mstr{min}\left(1-c_1, \frac{2(\mu_w -2 + 1/{\mu_w})}{(n+2)^2+\mu_w}\right)}$, 
					
											\lmma{m = 4 + \lfloor 3 \ln \, n  \rfloor, c_{\sigma} = \frac{2 \lambda}{n}, c_{d,i} = \frac{1}{1.5^{i-1} n},  c_{c,i} = \frac{\lambda}{ 4^{i-1} n} \; \mstr{for} \; i=1, \ldots, m}
											} \label{MAEScmaGiven}
\STATE{\textbf{initialize} $t \leftarrow 0, \vc{y}^{(t=0)} \in \R^{\dd}, \sigma^{(t=0)} > 0, \vc{p}^{(t=0)}_{\sigma} = \ma{0}$, \cma{\vc{p}^{(t=0)}_{c} = \ma{0}, \C^{(t=0)} = \I}, \maes{\M^{(t=0)} = \I}, \lmma{\vc{m}^{(t=0)}_i \in \R^{\dd }, \vc{m}^{(t=0)}_{i} = \ma{0} \; \mstr{for} \; i=1, \ldots, m}}
%
\REPEAT
  \FOR{$i \leftarrow 1,\ldots,\lambda$} \label{MAGenerateBegin}
				\STATE{$\vc{z}^{(t)}_i \leftarrow {{\mathcal N}  \hspace{-0.13em}\left({\ma{0},\I\,}\right)}$}
				\STATE{$\vc{d}^{(t)}_i \leftarrow \vc{z}^{(t)}_i$}
                \STATE{\cmatext{\textbf{if} $t \mod \frac{n}{\lambda} = 0$ \textbf{then} $\vc{\M}^{(t)} \leftarrow \sqrt{\C^{(t)}}$ \textbf{else} $\vc{\M}^{(t)} \leftarrow\vc{\M}^{(t-1)}$}} \label{CMAdi} \COMMENT{CMA-ES}
				\STATE{\maes{\vc{d}^{(t)}_i \leftarrow \M^{(t)} \; \vc{d}^{(t)}_i}} \COMMENT{CMA-ES and MA-ES}
				\STATE{\lmmatext{\textbf{for} $j \leftarrow	 1,\ldots,\mstr{min}(t,m) \; \textbf{do}$}} \label{LMloop1}	\COMMENT{LM-MA-ES}
				\STATE{$\;\;\;\;$ \lmma{\vc{d}^{(t)}_i \leftarrow (1 - c_{d,j}) \vc{d}^{(t)}_i + c_{d,j} \vc{m}^{(t)}_{j}\left((\vc{m}^{(t)}_{j})^T \vc{d}^{(t)}_i  \right)}}		 \label{LMloop2}	\COMMENT{LM-MA-ES}
			\STATE{ $\vc{f}^{(t)}_i \leftarrow f(\vc{y}^{(t)} + \sigma^{(t)} \vc{d}^{(t)}_i)$} \label{GenerateEnd}
  \ENDFOR
%
	\STATE{ $ \vc{y}^{(t+1)} \leftarrow  \vc{y}^{(t)} + \sigma^{(t)} \sum_{i=1}^{\mu} w_i \vc{d}^{(t)}_{i:\lambda} \;$} \COMMENT{the symbol $i:\lambda$ denotes $i$-th best sample on $f$} \label{ComputeNewMean}
	\STATE{ $ \vc{p}^{(t+1)}_{\sigma} \leftarrow (1 - c_{\sigma}) \vc{p}^{(t)}_{\sigma} + \sqrt{\mu_w c_{\sigma}(2-c_{\sigma})} \sum_{i=1}^{\mu} w_i \vc{z}^{(t)}_{i:\lambda} $} \label{SigmaPathUpdate} 
	\STATE{ \cma{\vc{p}^{(t+1)}_{c} \leftarrow (1 - c_{c}) \vc{p}^{(t)}_{c} + \sqrt{\mu_w c_{c}(2-c_{c})} \sum_{i=1}^{\mu} w_i \vc{d}^{(t)}_{i:\lambda} }} \COMMENT{CMA-ES}
\STATE{ \cma{\C^{(t+1)} \leftarrow (1 - c_1 - c_{\mu}) \C^{(t)} +
						c_1 \vc{p}_c ({\vc{p}^{(t)}_c})^T + 
						c_{\mu} \sum^{\mu}_{i=1} w_i \vc{d}^{(t)}_{i:\lambda} (\vc{d}^{(t)}_{i:\lambda})^T }}		\label{cmaCupdate} \COMMENT{CMA-ES}				
	\STATE{ \maes{ \M^{(t+1)} \leftarrow \M^{(t)} \left[ \I + \frac{c_{1}}{2} \left(\vc{p}^{(t)}_{\sigma}(\vc{p}^{(t)}_{\sigma})^T - \I \right) + \frac{c_{\mu}}{2} \left(\sum_{i=1}^{\mu} w_i \vc{z}^{(t)}_{i:\lambda}(\vc{z}^{(t)}_{i:\lambda})^T - \I \right) \right]} } \COMMENT{MA-ES} \label{MAupdate}
    \STATE{\lmmatext{\textbf{for} $i \leftarrow 1,\ldots,m$\; \textbf{do}}} \label{LMMAupdate1} \COMMENT{LM-MA-ES} 
    \STATE{$\;\;\;\;$ \lmma{\vc{m}^{(t+1)}_i \leftarrow (1-c_{c,i})\vc{m}^{(t)}_i + \sqrt{\mu_w c_{c,i}(2-c_{c,i})}  \sum_{j=1}^{\mu} w_j \vc{z}^{(t)}_{j:\lambda}}} \label{LMMAupdate2} \COMMENT{LM-MA-ES} 
	\STATE{ $ \sigma^{(t+1)} \leftarrow \sigma^{(t)} \cdot \mstr{exp}
	          \left[ \frac{c_{\sigma}}{2} \left(  \frac{\left\| \vc{p}^{(t+1)}_{\sigma} \right\|^2}{ n} - 1  \right) \right] $} \label{MAStepSizeUpdate} 
  \STATE{ $ t \leftarrow t + 1$}
\UNTIL{ \textit{stopping criterion is met} }
\end{algorithmic}
\end{algorithm}

\section{The Limited-Memory Matrix Adaptation Evolution Strategy}
\label{secLMMAES}

A number of methods was proposed to reduce the space and time complexity per sample from $\mathcal{O}(n^2)$ to $\mathcal{O}(n)$ or at least $\mathcal{O}\big(n\log(n)\big)$ while still modeling the most relevant aspects of the full covariance matrix. Simple approaches like \citep{ros2008simple} restrict the covariance matrix to its diagonal, while more elaborate methods use a low-rank approach \citep{sun2011linear,loshchilov2017lm}. Both approaches can be combined~\citep{akimoto2014comparison}. 
Inspired by the Limited-Memory CMA-ES \citep{loshchilov2017lm} which in turn is inspired by the L-BFGS method \citep{liu1989limited}, we show how to scale up MA-ES to high-dimensional problems. The derivation given below is based on the multiplicative update, the final result for the additive update \eqref{MAsampling12} is equivalent when  $\vc{d}^{(t)}_{\sigma}$ is not stored but reconstructed as $\M^{(t)} \vc{p}^{(t)}_{\sigma}$. 
At iteration $t$, the main update equation of MA-ES reads 
\begin{equation}
\label{Mupdate}
\M^{(t+1)} \leftarrow \M^{(t)} \left[ \I + \frac{c_{1}}{2} \left(\vc{p}^{(t+1)}_{\sigma}(\vc{p}^{(t+1)}_{\sigma})^T - \I \right) + \frac{c_{\mu}}{2} \left(\sum_{i=1}^{\mu} \vc{w}_i \vc{z}^{(t)}_{i:\lambda}(\vc{z}^{(t)}_{i:\lambda})^T - \I \right) \right],
\end{equation}
where $\M^{(t)}$ is adapted multiplicatively based on the rank-one update weighted by 
$\frac{c_{1}}{2}$
and the rank-$\mu$ update weighted by 
$\frac{c_{\mu}}{2}$,
starting from $\M^{(t=0)} = \Id$.
By omitting the rank-$\mu$ update for the sake of simplicity (i.e., by setting $c_{\mu}=0$), 
we obtain 
\begin{equation}
\label{MupdateT1}
\M^{(1)} \leftarrow \I + \frac{c_{1}}{2} \left(\vc{p}^{(1)}_{\sigma}(\vc{p}^{(1)}_{\sigma})^T - \I \right)  = \left(1 - \frac{c_{1}}{2} \right) \I + \frac{c_{1}}{2} \vc{p}^{(1)}_{\sigma}(\vc{p}^{(1)}_{\sigma})^T 
\end{equation}
The sampling procedure of the $i$-th solution $\vc{x}^{(1)}_i$ follows 
\begin{equation} 
\label{ST1}
\vc{x}^{(1)}_i \leftarrow \vc{y}^{(1)} + \sigma^{(1)} \vc{d}^{(1)}_i = \vc{y}^{(1)} + \sigma^{(1)} \M^{(1)} \vc{z}^{(1)}_i,
\end{equation}
where $\vc{z}_i^{(1)} \sim {{\mathcal N}  \hspace{-0.13em}\left({\ma{0},\I\,}\right)}$. 
One can rewrite $\vc{d}_i^{(1)} = \M^{(1)} \vc{z}_i^{(1)}$ based on equation~\eqref{MupdateT1} as 
\begin{equation} 
\label{SamplingT1}
\vc{d}_i^{(1)}=\M^{(1)} \vc{z}^{(1)}_i = \left( \left(1 - \frac{c_{1}}{2}\right) \I + \frac{c_{1}}{2} \vc{p}^{(1)}_{\sigma}(\vc{p}^{(1)}_{\sigma})^T\right) \vc{z}_i^{(1)} =
\vc{z}^{(1)}_i \left(1 - \frac{c_{1}}{2}\right) + \frac{c_{1}}{2} \vc{p}^{(1)}_{\sigma} \left((\vc{p}^{(1)}_{\sigma})^T \vc{z}^{(1)}_i\right)
\end{equation}

Importantly, $\left((\vc{p}^{(1)}_{\sigma})^T \vc{z}^{(1)}\right)$ is a scalar (see line \ref{LMloop2} in Algorithm \ref{ALLALGOS})  and thus equation~\eqref{SamplingT1} does not require $\M^{(1)}$ to be stored in memory. 
%
%
One generally obtains 
\begin{equation} 
\label{SamplingTt}
\vc{d}^{(t)}_i  = \M^{(t)} \vc{z}_i^{(t)} = \M^{(t-1)} \ma{P}^{(t)} \vc{z}_i^{(t)}
= \M^{(t-1)}
\underbrace{\left( \left(1 - \frac{c_{1}}{2}\right) \I + \frac{c_{1}}{2} \vc{p}^{(t)}_{\sigma}(\vc{p}^{(t)}_{\sigma})^T\right)}_{:=\ma{P}^{(t)}}  \vc{z}_i^{(t)}
\end{equation}
leading to a sequence of products
\begin{eqnarray} 
\label{dti}
\vc{d}^{(t)}_i 
 & \!\!\!=\!\!\! & \left( \left(1 - \frac{c_{1}}{2}\right) \I + \frac{c_{1}}{2} \vc{p}^{(1)}_{\sigma}(\vc{p}^{(1)}_{\sigma})^T\right)  \cdot \ldots
 \nonumber \\
 & & \ldots \cdot 
 \left( \left(1 - \frac{c_{1}}{2}\right) \I + \frac{c_{1}}{2} \vc{p}^{(t-1)}_{\sigma}(\vc{p}^{(t-1)}_{\sigma})^T\right) \cdot \left( \left(1 - \frac{c_{1}}{2}\right) \I + \frac{c_{1}}{2} \vc{p}^{(t)}_{\sigma}(\vc{p}^{(t)}_{\sigma})^T\right)  \vc{z}_i^{(t)}
\end{eqnarray}
which is to be treated from right to left. Thus, the sampling procedure for $\vc{d}^{(t)}_i = \M^{(t)} \vc{z}_i^{(t)}$ does neither require matrix-matrix-product operations nor does it require the storage of $\M^{(t)}\in \R^{\dd \times \dd}$, but can be performed based on $t$ vectors $\vc{p}^{(t)}_{\sigma}$ used to construct $\M^{(t)}$. However, this is efficient only for $t \ll n$. Therefore, in order to reduce the cost of the sampling procedure, equation~\eqref{dti} must be \emph{approximated} in one way or another by artificially limiting the number $m$ of supporting $\vc{p}^{(t)}_{\sigma}$ vectors such that $m \ll n$. In this work we pick $m \in \mathcal{O}(\log(n))$.

LM-CMA-ES\citep{loshchilov2017lm} addresses a similar problem of compactly representing the covariance matrix with $m \in \mathcal{O}\big(\log(n)\big)$ direction vectors: instead of considering the last $m$ vectors, it samples them in a certain temporal distance in terms of iterations $t$.
The same approach works for LM-MA-ES (see section 1 and Algorithm 1 in the supplementary material). However, the rather complicated procedure of ensuring a temporal distance between $\vc{p}_{\sigma}$ vectors can be simplified by considering different time horizons of their update. This procedure is a viable alternative since $\vc{p}_{\sigma}$ itself is anyway incrementally updated with  $\sum_{i=1}^{\mu} w_i \vc{z}_{i:\lambda}$. Thus, instead of a full transformation matrix $\M \in \R^{n \times n}$, LM-MA-ES maintains $m \in \mathcal{O}\big(\log(n)\big)$ vectors $\vc{m}_i$ (see lines \ref{LMMAupdate1}-\ref{LMMAupdate2} in Algorithm \ref{ALLALGOS}), modeling the deviation of the transformation matrix from the identity as a rank-$m$ matrix. The learning rates $c_{c,i}$ and $c_{d,i}$ for applying and updating the vectors $\vc{m}_i$ are chosen to be exponentially decaying, hence the $\vc{m}_i$ are fading records of mean update steps on exponentially differing time scales. This is in contrast to CMA-ES and MA-ES, which update their matrices $\C$ and $\M$ only with two different learning rates for the rank-1 and rank-$\mu$ updates, and hence operate on a single time scale. LM-MA-ES learns some directions very quickly, while others are kept more stable. This can be advantageous in particular in high dimensions where learning rates are generally small due to the sub-linear sample size.

The proposed LM-MA-ES method features all invariance properties of modern ESs, namely invariance to translation and rotation, as well as invariance to strictly monotonic (rank-preserving) transformations of objective values.


\section{Experimental validation}
\label{secExperiments}


\newcommand{\sphere}{$f_{Sphere}$}
\newcommand{\rosen}{$f_{Rosen}$}
\newcommand{\cigar}{$f_{Cigar}$}
\newcommand{\discus}{$f_{Discus}$}
\newcommand{\elli}{$f_{Elli}$}
\newcommand{\diffpow}{$f_{DiffPow}$}
\newcommand{\rotrosen}{$f_{RotRosen}$}
\newcommand{\rotcigar}{$f_{RotCigar}$}
\newcommand{\rotdiscus}{$f_{RotDiscus}$}
\newcommand{\rotelli}{$f_{RotElli}$}
\newcommand{\rotdiffpow}{$f_{RotDiffPow}$}

With our experimental evaluation we aim to answer the following questions:
\begin{compactitem}
\item
	How does LM-MA-ES compare to MA-ES, i.e., what is the effect of modeling only a $\mathcal{O}(\log(n))$ dimensional subspace?
\item
	How does LM-MA-ES compare to other algorithms designed for high-dimensional black-box optimization?
\item
	Is LM-MA-ES suitable for solving problems in machine learning?
\end{compactitem}
To answer these questions, we investigate the performance on standard benchmark problems of varying dimension, and we generate adversarial inputs for a random forest classifier.

\begin{minipage}{\textwidth}
  \begin{minipage}[b]{0.64\textwidth}
    \scriptsize
    \begin{tabular}{ll}
      \hline  
Name & Function $f(\vc{x})$\\ 
\hline
Sphere & $\sum_{i=1}^n \vc{x}^{2}_{i}$ \\ [0.10cm]
Ellipsoid &$\sum_{i=1}^n 10^{6{\frac{i-1}{n-1}}} \vc{x}^{2}_{i}$ \\ [0.10cm]
Rosenbrock & $\sum_{i=1}^{n-1}\left(100 \cdot (\vc{x}^{2}_{i}-x_{i+1})^{2}+(x_{i}-1)^{2}\right)$ \\[0.10cm]
Discus & $10^6\vc{x}^2_1 + \sum_{i=2}^n \vc{x}^{2}_{i}$ \\[0.10cm]
Cigar & $\vc{x}^2_1 + 10^6\sum_{i=2}^n \vc{x}^{2}_{i}$ \\[0.10cm]
Different Powers & $\sum_{i=1}^n \left|\vc{x}_i\right|^{2+4(i-1)/(n-1)}$ \\[0.10cm]
\hline
      \end{tabular}
      \captionof{table}{\label{table:TableFunc} Test functions used in this study.}
    \end{minipage}
    \begin{minipage}[b]{0.35\textwidth}
    \centering
   \includegraphics[width=\textwidth]{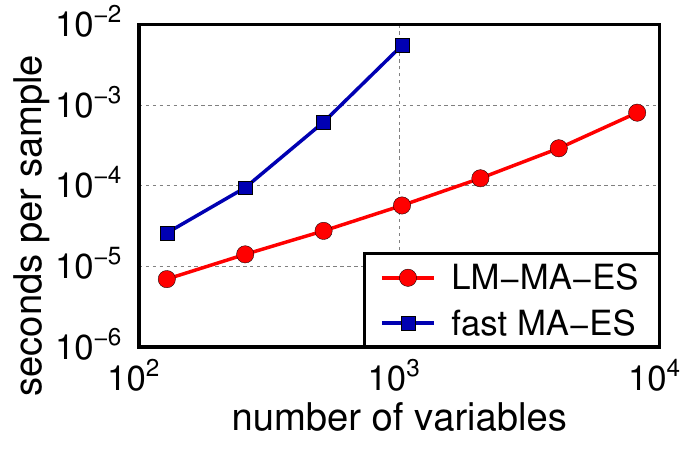}
    \captionof{figure}{\label{fig:runtime} Internal algorithm cost.}
  \end{minipage}
\end{minipage}

\subsection{Performance on Benchmark Problems}

We experimentally validate the proposed LM-MA-ES algorithm on large-scale variants ($n \in \{128, 256, \ldots, 8192\}$) of well established benchmark problems \citep{2009FinckBBOB2009setup} (see Table~1). Starting from the initial region $[-5, 5]^n$ containing the optimum with initial step size $\sigma = 3$ we optimize until reaching the (rather exact) target precision of $f_{tar}=10^{-10}$.


All hyperparameters of LM-MA-ES and MA-ES are given in Algorithm \ref{ALLALGOS}. 
We use LM-CMA-ES \citep{loshchilov2017lm}, VD-CMA-ES \citep{akimoto2014comparison} and the active ($\mu/\mu_{w},\lambda$)-CMA-ES \citep{2010HansenBBOBActiveCMA,2006ArnoldActiveCMA} (aCMA-ES, known to be up to 2 times more efficient than the default CMA-ES) as baselines. The source code of LM-MA-ES is available
in the supplementary material.


LM-MA-ES does not show performance degradation when applied to translated and rotated test functions (see also Figure 1 in the supplementary material).
Figure~\ref{fig:runtime} shows the effect of the $\mathcal{O}\big(n\log(n)\big)$ scaling of the runtime per sample as compared to $\mathcal{O}(n^2)$ of fast-MA-ES (in the following denoted as MA-ES), measured for implementations of the algorithms in plain C.

\begin{figure}
\begin{center}
\includegraphics[width=\textwidth]{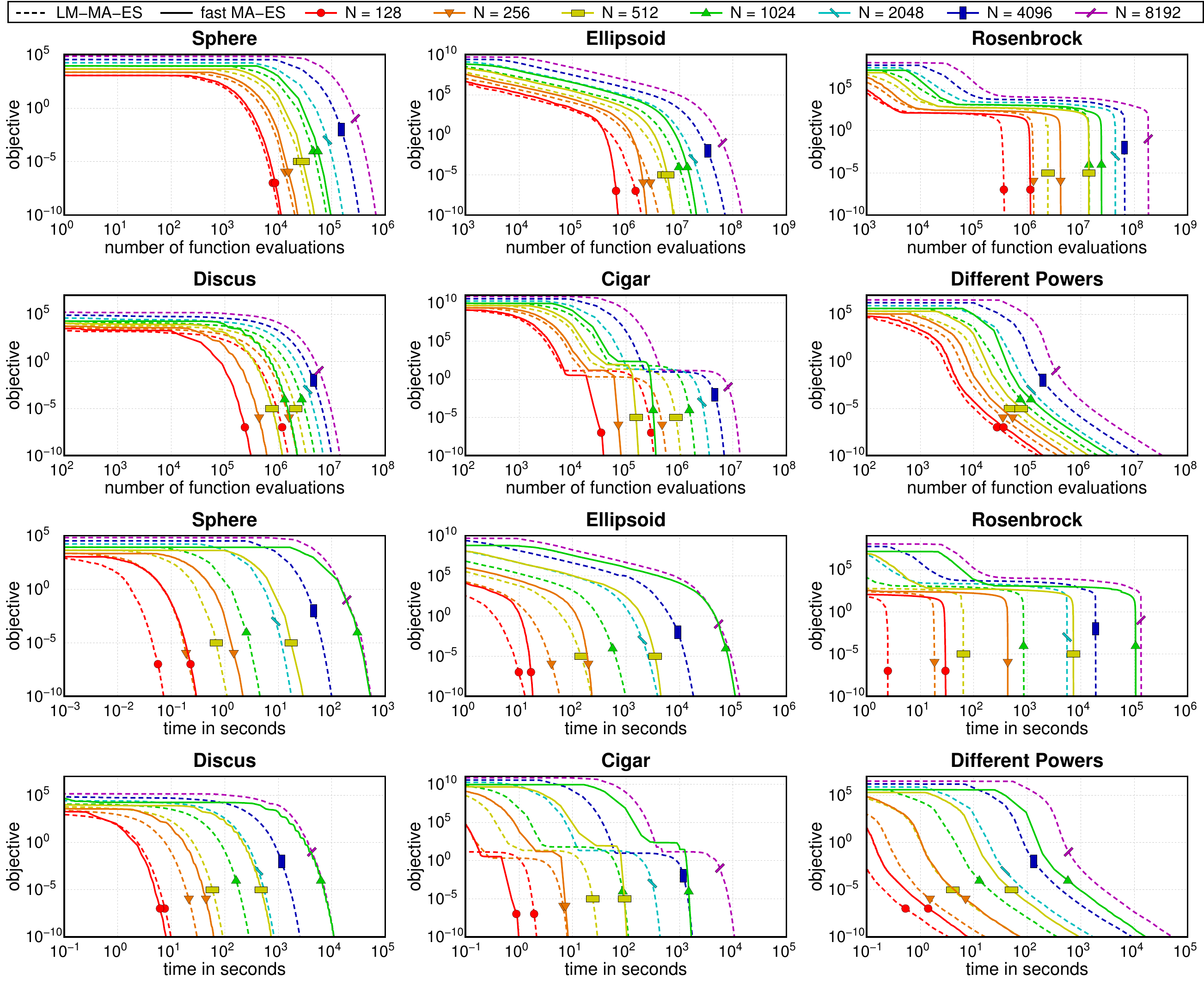}
\end{center}
\caption{
\label{fig:limited-vs-full}
Runtime in number of function evaluations (rows 1--2) and seconds (rows 3--4) of LM-MA-ES in 128 to 8192 dimensions and fast MA-ES in 128 to 1024 dimensions on six standard benchmark problems.
}
\end{figure}

The much better internal scaling is of value only if the algorithm does not pay a too high price in terms of an increased number of function evaluations required to reach $f_{tar}$. Therefore we test LM-MA-ES against MA-ES with full rank transformation matrix. Figure~\ref{fig:limited-vs-full} shows that LM-MA-ES performs surprisingly well: in many cases it is actually faster and this tends to happen more often for larger $n$. LM-MA-ES is always faster in terms of wall clock time, in some cases by a factor of 100.

\begin{figure}%
\includegraphics[width=0.33\textwidth]{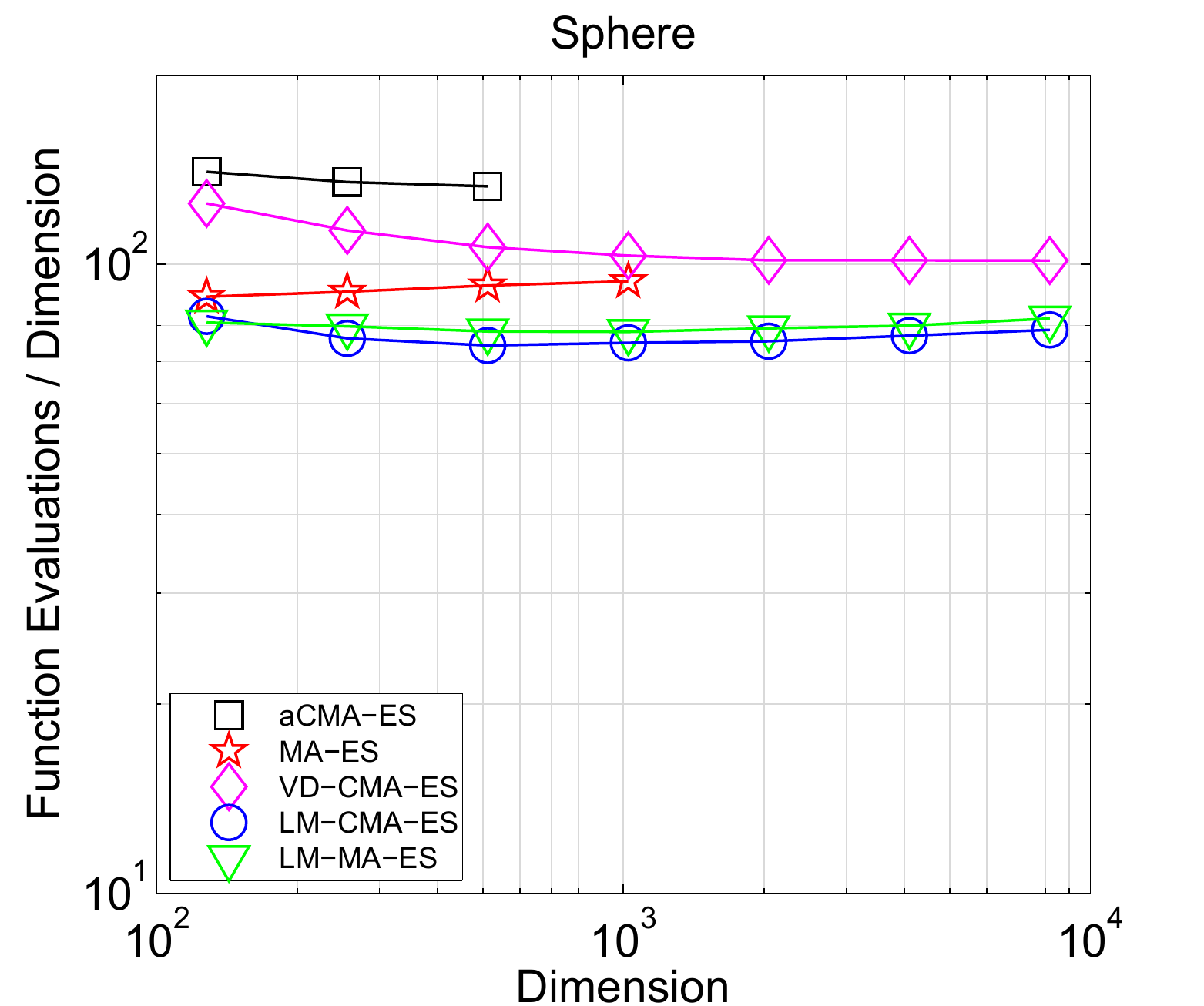}
\includegraphics[width=0.33\textwidth]{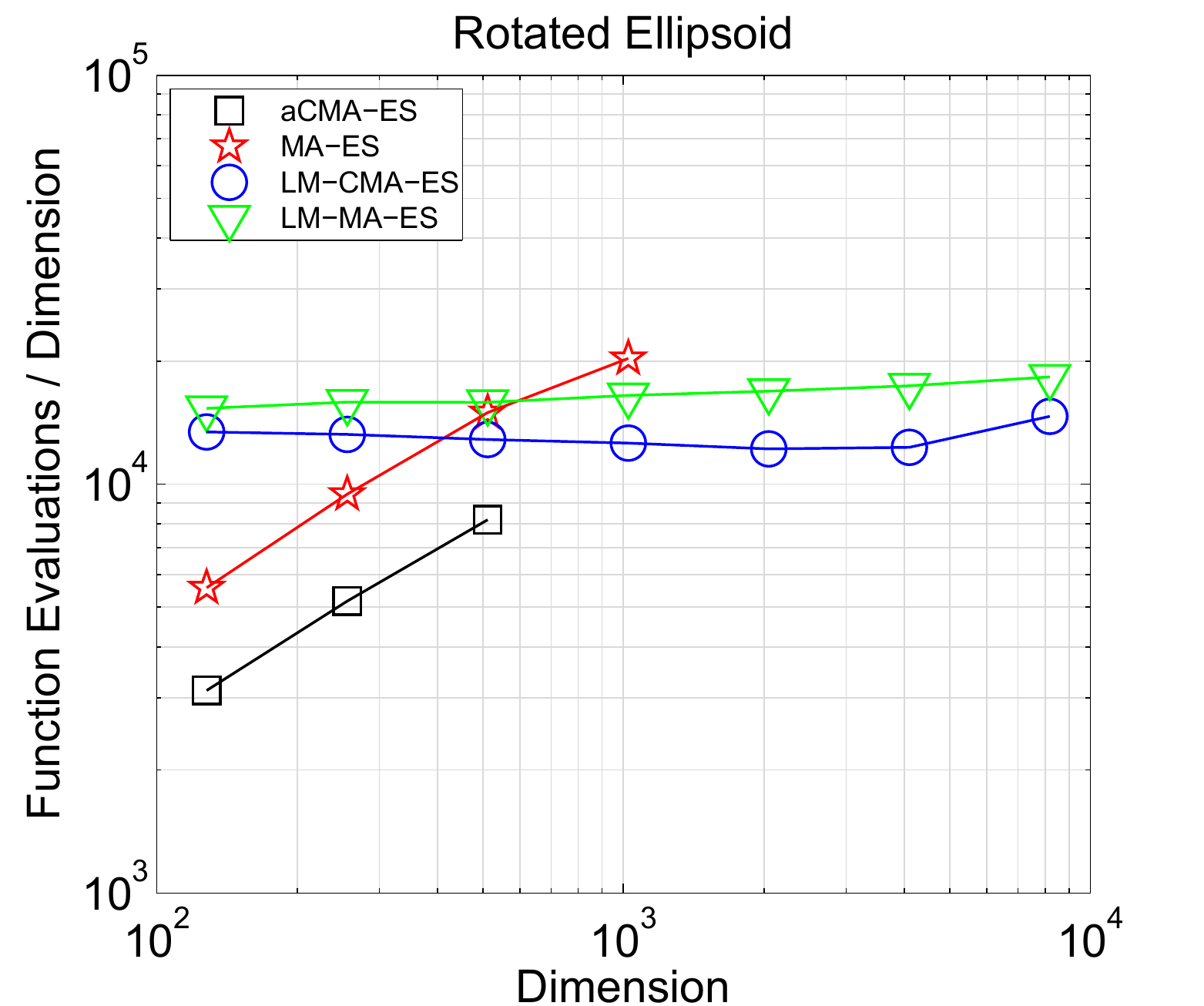}
\includegraphics[width=0.33\textwidth]{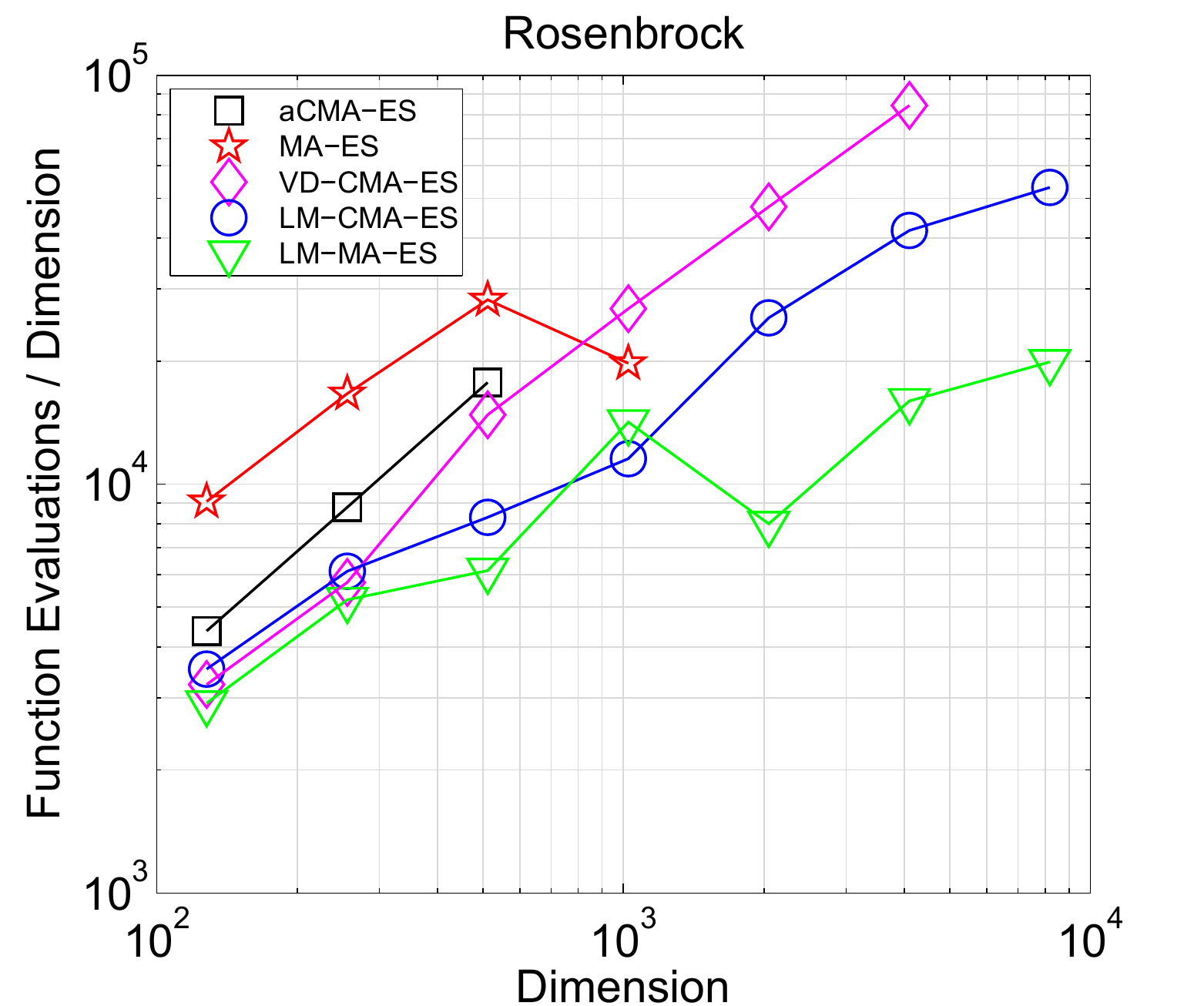}\\ 
\includegraphics[width=0.33\textwidth]{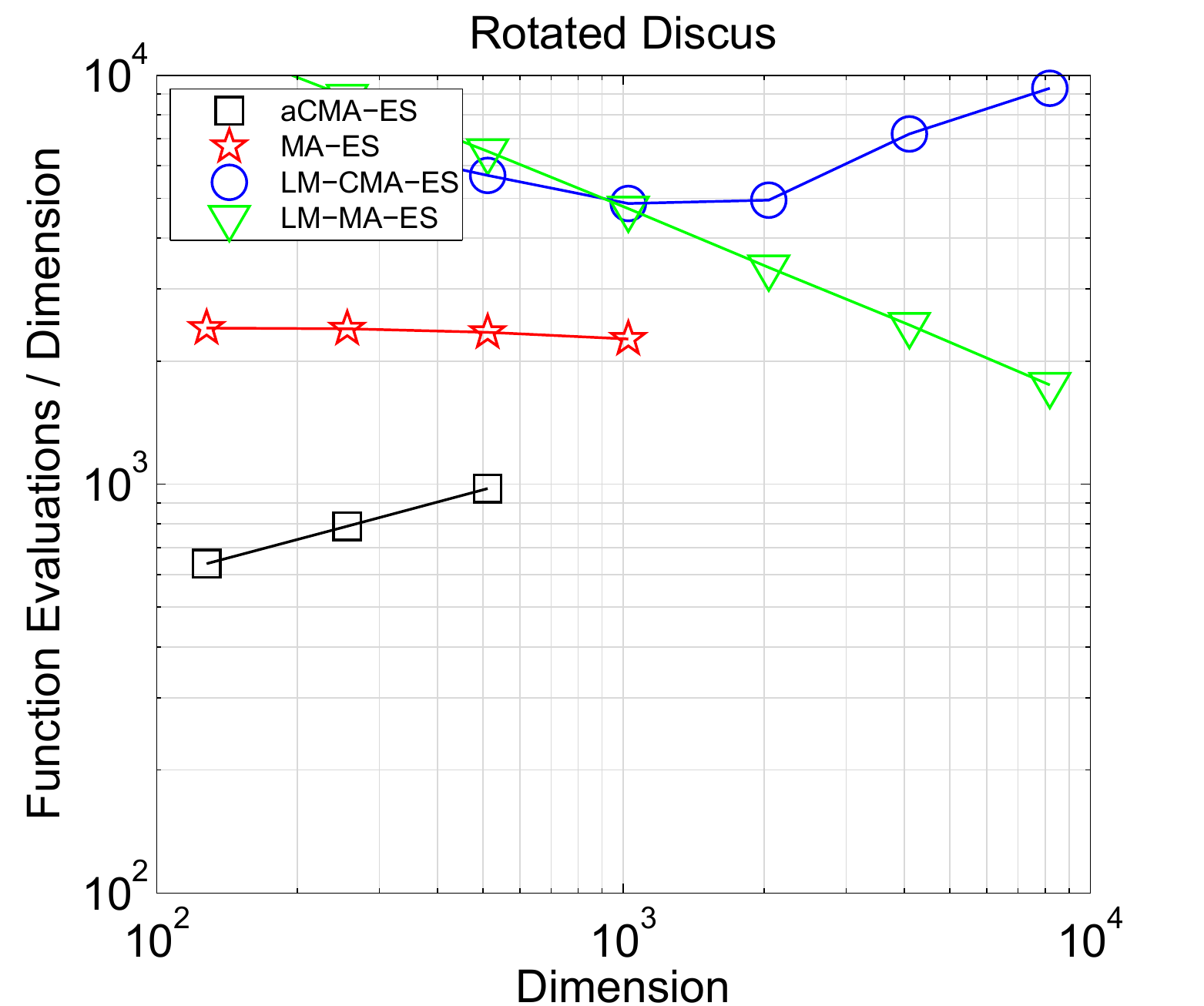}
\includegraphics[width=0.33\textwidth]{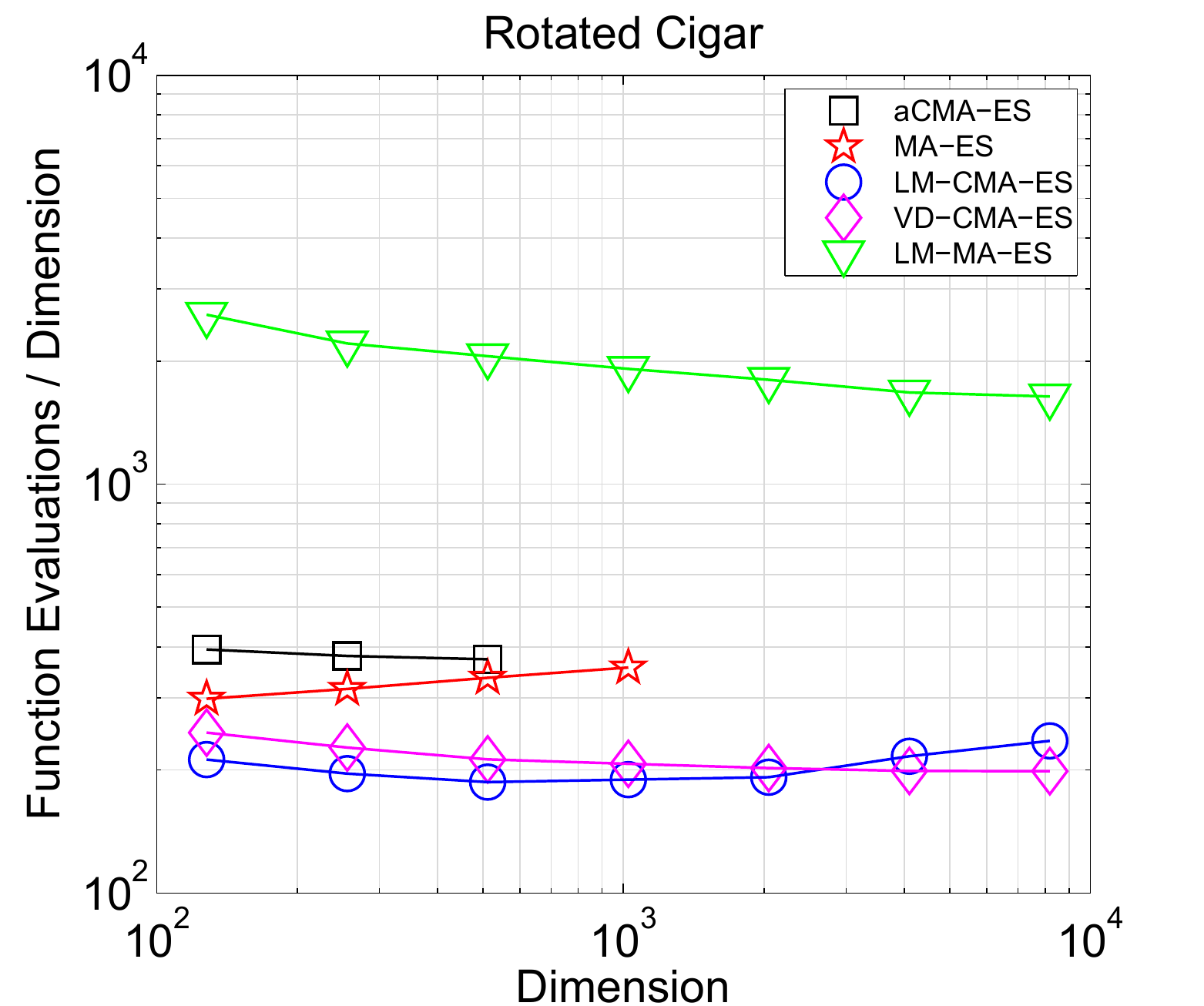}
\includegraphics[width=0.33\textwidth]{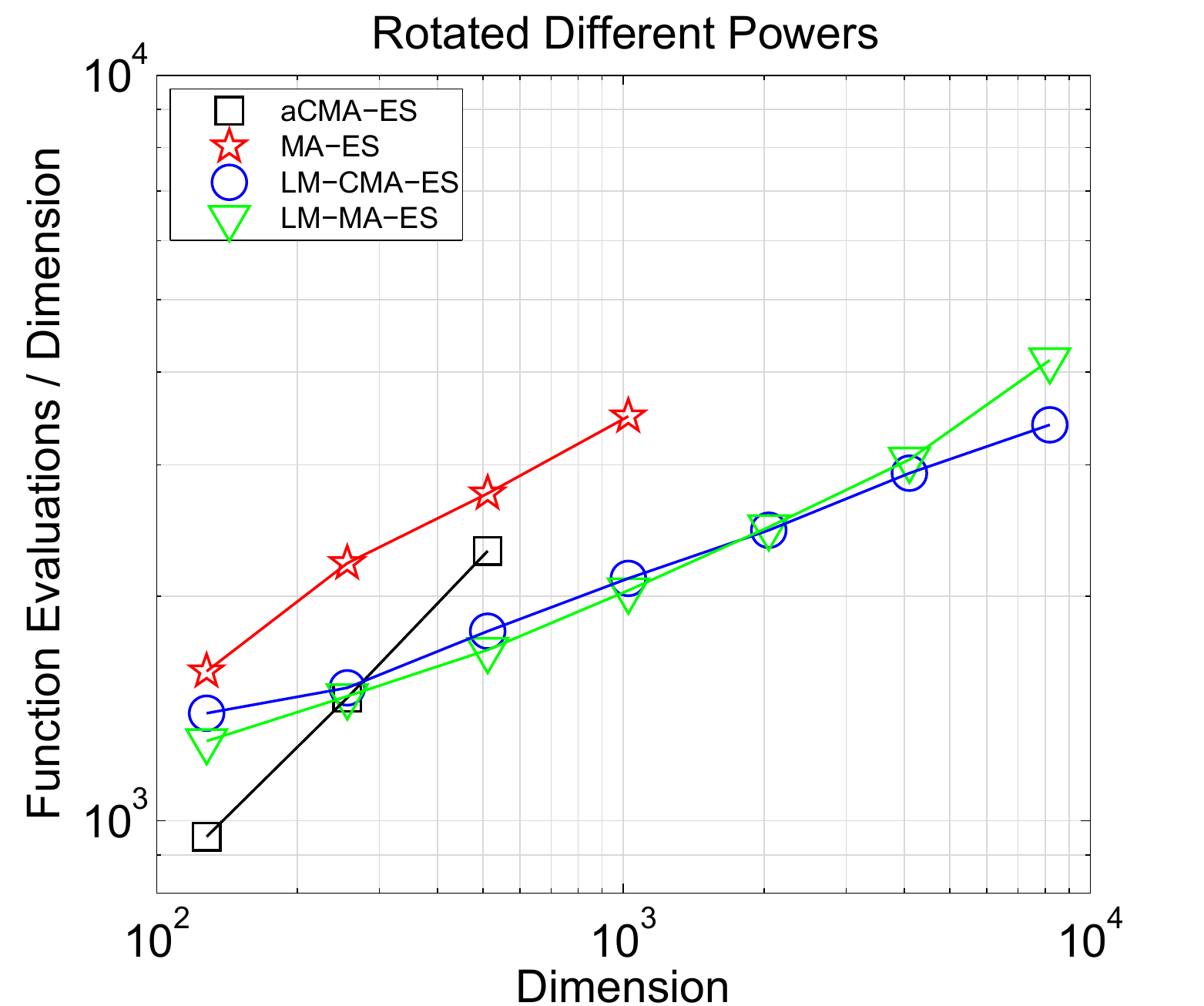}
\caption{
\label{fig:algorithm-comparison}
Median number of function evaluations (out of 5 runs) required to achieve a target objective function value of $f_{tar}=10^{-10}$. The results of VD-CMA-ES are shown only on  functions where the algorithm succeeded achieving $f_{tar}$. 
Some results for CMA-ES and MA-ES are missing due to their extremely long runtimes (note that the vertical axis shows function evaluations, not runtime).}
\end{figure}

Figure \ref{fig:algorithm-comparison} shows that LM-MA-ES scales favorably compared to LM-CMA-ES achieving better scaling on the Rosenbrock and Discus functions, but a worse scaling on Cigar. The latter result might be due to an improper setting of the hyperparameters, a problem that can most probably be fixed with the technique proposed in \citep{loshchilov2014maximum}. VD-CMA is not able to solve some rotated functions efficiently due to the restrictions on the covariance matrix that the algorithm assumes \citep{akimoto2014comparison,loshchilov2017lm}.

\subsection{Adversarial Inputs}

Several standard classifiers like the $k$-nearest-neighbor predictor (without distance-based weights), decision trees, and random forests are not differentiable with respect to their inputs. The predictions are piecewise constant and hence trivially piecewise differentiable, however, the gradient is zero and hence uninformative. This is a significant complication when generating adversarial inputs \citep{szegedy2013intriguing} for such classifiers, a problem for which only rather weak attacks exist \citep{kantchelian2016evasion,xu2016automatically}. We compare MA-ES and LM-MA-ES on this task. To this end we train a random forest consisting of 1000 trees on the MNIST data set (784 dimensional input, 10 classes, 60,000 training points, 10,000 test points). It achieves a test error of $97.21\%$, which is clearly worse than the results obtained with convolutional neural networks, however, the predictor is highly non-trivial in the sense of being far from guessing performance. Let $h$ denote the random forest predictor, let $\vc{x}_0$ denote a correctly classified test point (of which we have 9721), and let ${y}_0$ denote the corresponding label. For an input $\vc{x}$, the random forest predictor outputs a probability vector $h(\vc{x}) \in \mathbb{R}^{10}$. We create an adversarial version of $(\vc{x}_0, {y}_0$) by minimizing the objective function
\begin{align}
	\label{eq:fitness}
	f(\vc{x}) = \begin{cases}
			h(\vc{x})_{{y}_0} - \max\{(h(\vc{x})_i \,|\, i \not= y_0\} & \text{if $\vc{x}$ is classified as ${y}_0$} \\
            \frac{-1}{\|\vc{x} - \vc{x}_0\|} & \text{otherwise}
		\end{cases}
\end{align}
starting from $\vc{x} = \vc{x}_0$ with initial step size $\sigma = 1$, where an MNIST digit is encoded as a vector $\vc{x} \in \mathbb{R}^{784}$, with gray values in the interval $[0, 255]$.
The case distinction encodes a preference for wrongly classified points: correctly classified points have a positive value, wrongly classified points have a (better) negative value. For positive values the incentive is to reduce the difference between the fraction of trees voting for the correct label $\vc{y}_0$ and the strongest alternative. For a wrongly classified point the continuous trend moves the point back as close as possible to the original image, in terms of Euclidean distance. Note that due to invariance under rank-preserving transformations, the second term is exactly equivalent to minimization of the distance between $\vc{x}$ and~$\vc{x}_0$.

\begin{wrapfigure}{r}{.5\textwidth}
\includegraphics[width=0.16\textwidth]{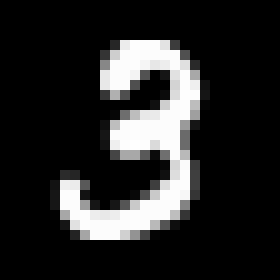}~~\includegraphics[width=0.16\textwidth]{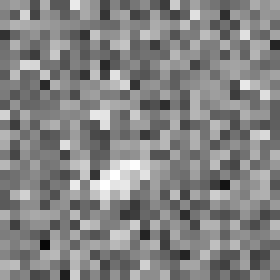}~~\includegraphics[width=0.16\textwidth]{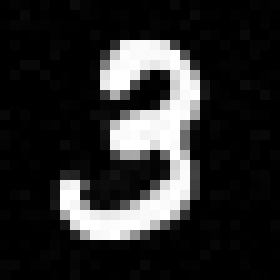}
\caption{
\label{figure:MNIST}
Original image (left) and corresponding adversarial image (right), created through a run of LM-MA-ES. The images are visually indistinguishable. The (boosted) difference image is shown in the center; mid gray corresponds to a difference of zero.\vspace*{-1.0em}
}
\end{wrapfigure}

Each of the 9721 optimization runs was restricted to a very low budget of 1000 objective function evaluations. This seems reasonable from a security perspective, since querying the classifier too frequently renders a remote attack inefficient.
We used the sklearn implementation of the random forest \citep{scikit-learn} and implementations of MA-ES and LM-MA-ES based on numpy. The overall experiment takes about two hours for LM-MA-ES and about 7 hours for MA-ES on a laptop. The optimizers managed to turn 6152 (MA-ES) and 6321 (LM-MA-ES) images into wrongly classified inputs. In all cases, they were visually indistinguishable from the original test inputs (see Figure~\ref{figure:MNIST}).
The surprisingly high rate of about 65\% of correctly classified test inputs that were successfully turned into adversarial versions demonstrates that random forests can be highly unstable and rather easy to fool by an adversarial.

From an optimization point of view, LM-MA-ES performed clearly better than plain MA-ES. It ran about 3.5 times faster and even produced better results: in 7171 cases LM-MA-ES reached a lower objective value, while MA-ES was better in only 2550 out of 9721 cases. This is also reflected by the larger number of cases in which an adversarial input was found (6321 vs.\ 6152, see above).

\begin{wrapfigure}{r}{.5\textwidth}
\includegraphics[width=0.5\textwidth]{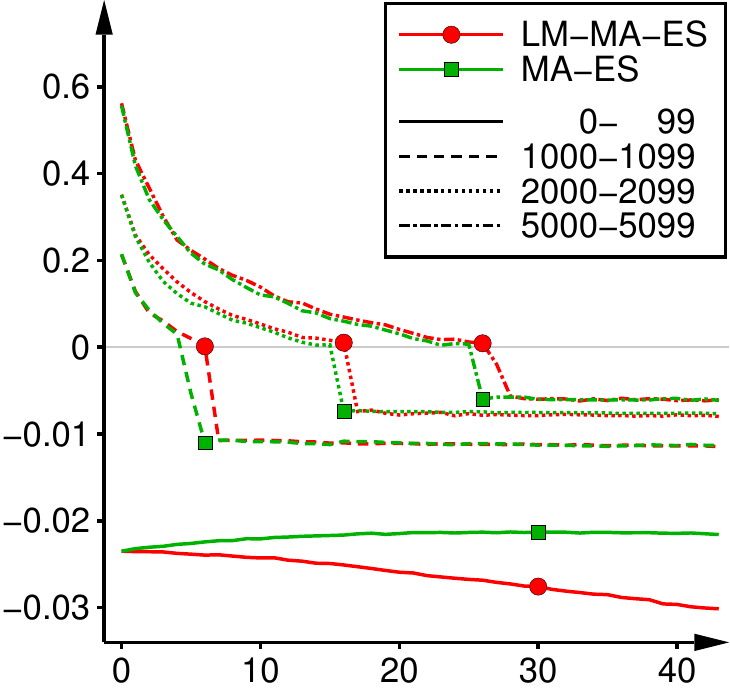}
\caption{\label{figure:progress}
Progress over 1000 function evaluations (44 iterations) of LM-MA-ES and MA-ES on the task of generating adversarial inputs. The figure shows four objective function curves corresponding to equation~\eqref{eq:fitness} for each algorithm, referring to medians over 100 runs each. The 9721 problems were sorted by their initial objective value. We consider problems 0 to 99 (the 100 simplest problems), problems 1000 to 1099, 2000 to 2099, and 5000 to 5099.\vspace*{-2em}}
\end{wrapfigure}

Optimization progress over time is plotted in Figure~\ref{figure:progress} for four subsets of runs of varying difficulty. In all cases, LM-MA-ES is slightly faster. Significant differences can be observed only for the easiest problems, where MA-ES fails to improve the solution, presumably due to the small budget of objective function evaluations. In a control experiment we verified that the performance of MA-ES is only insignificantly improved compares to a simple ES without matrix adaptation, which is restricted to the adaptation of the global step size.

The better quality of the results given the same number of objective function evaluations can be explained by the low-rank structure of the LM-MA-ES steps. Adapting a full $784 \times 784$ transformation matrix with only 1000 function evaluations is a hopeless undertaking, while the top $23 = \lfloor 4 + 3 \log(784) \rfloor$ directions are much easier to adapt. This difference allows LM-MA-ES to use learning rates that are up to several orders of magnitude larger than for MA-ES, resulting in faster adaptation to the problem at hand.
From the difference image in Figure~\ref{figure:MNIST} we clearly see a bright blob, indicating that LM-MA-ES has identified a subspace with a meaningful interpretation for the problem at hand. In contrast, the plain ES without matrix adaptation applies only white noise distortions, and this is also what's encoded by the initial search distribution of MA-ES and LM-MA-ES.

Returning to our initial questions we conclude that LM-MA-ES successfully marries the simplicity of the update mechanisms of MA-ES with the graceful scaling of large-scale optimizers like LM-CMA-ES to large~$n$. LM-MA-ES is competitive with MA-ES in terms of the required number of objective function evaluations, while reducing the algorithm internal cost per sample considerably from $\mathcal{O}\big(n^2\big)$ down to $\mathcal{O}\big(n \log(n) \big)$. Taken together this yields in significant speed-ups for high-dimensional problems. By applying our algorithm to the problem of generating adversarial inputs for a random forest classifier, we demonstrate the value of LM-MA-ES for this domain, and uncover a surprising vulnerability of random forests with respect to the existence of adversarial inputs.

\section{Conclusion}

The recently proposed Matrix Adaptation Evolution Strategy is a simpler variant of the Covariance Matrix Adaptation Evolution Strategy. We showed that the $O(n^2)$ time and space complexity of MA-ES, which is prohibiting for large $n$, can be reduced to $O\big(n\log(n)\big)$ adopting the approach used in~\citep{loshchilov2017lm}. The proposed Limited-Memory Matrix Adaptation Evolution Strategy matches the state-of-the-art results on large-scale optimization problems while being algorithmically simpler than LM-CMA-ES.

Future work should investigate to which extent the inclusion of the rank-$\mu$ update can improve the performance.
The learning rates of LM-MA-ES can be optimized online as it is commonly done in self-adaptive evolutionary algorithms \citep{beyer2001self} or based on the maximum-likelihood principle~\citep{loshchilov2014maximum}.

A promising venue for LM-MA-ES would be to accelerate Stochastic Gradient Descent (SGD) for training deep neural networks by replacing the evolution path vectors by momentum vectors based on noisy batch gradients. The method could potentially represent an alternative to L-BFGS and numerous SGD variants with adaptive learning rates.

\bibliographystyle{plain}

\cleardoublepage
\setcounter{page}{1}

\textbf{Supplementary Material} for the paper \textbf{Limited-Memory Matrix Adaptation\\for Large Scale Black-box Optimization}

\setcounter{section}{0}
\renewcommand{\thesection}{\Roman{section}}

\section{LM-MA-ES with the same procedure to store  direction vectors as in LM-CMA-ES}
\label{oldLMMAsection}

As it is mentioned in the main text of the paper, our original approach to design LM-MA-ES was to employ the same  procedure to storage direction vectors as used in LM-CMA-ES  \citep{loshchilov2017lm}. The corresponding variant of LM-MA-ES with $O(m_{\max} n)$ time and space complexity is given below in Algorithm \ref{oldLMMA}. 

Instead of $\M^{(t)}\in \R^{\dd \times \dd}$ used in the original MA-ES, we employ a storage matrix $\M^{(t)} \in \R^{m_{\max} \times \dd }$ whose lines store the evolution path vectors computed at different iterations of the algorithm. While the original MA-ES employs $\vc{p}^{(t)}_{\sigma}$ to adapt both $\sigma^{(t)}$ and $\M^{(t)}$, we note that the optimal time horizon of the two processes can be different. Therefore,  we separately adapt $\vc{p}^{(t)}_{\sigma}$ with the learning rate $c_{\sigma} = \frac{1}{m_{\max}}$ and $\vc{p}^{(t)}_{c}$ with the learning rate $c_{c} = \frac{1}{n}$ (see lines \ref{oldMASigmaPathUpdate} and \ref{oldMASigmaPathUpdate2}, respectively). 
At the first $m_{\max}$ iterations of the algorithm, LM-MA works as MA-ES without the rank-$\mu$ update, i.e., with $c_{\mu}=0$, and a different setting for hyperparameters, e.g., $c_c \neq c_{\sigma}$. The equivalence is due to the use of the first $m_{\max}$ evolution path vectors $\vc{p}^{(t)}_c$ to reproduce $\M^{(t=m_{\max})}$ based on the the rank-one update exactly. 
When $m_{\max} < n$, one can greatly reduce the time and space complexity of the final algorithm. 
However, the use of the most recent $m_{\max}$ evolution path vectors would lead to a degenerative sampling. 
Therefore, following \citep{loshchilov2017lm}, we adopt a reference array $\vc{ref}$ (thought of as a matrix, storing one vector per row) to access $\vc{p}^{(t)}_c$ vectors in $\M^{(t)}$
such that the $i$-th line of $\M^{(t)}$, i.e., $\M^{(t)}_{\vc{ref}_{i}}$, belongs to the $i$-th oldest vector (in terms of its iteration index $t$) among the stored ones (see line \ref{oldupdateM}). 
In order to well approximate the effect that would be obtained with the full matrix based on the rank-one update, we force the storage to support a temporal distance between the evolution path vectors such that the temporal distance $\vc{time}_{\vc{ref}_{i+1}}-\vc{time}_{\vc{ref}_i}$ between $i+1$-th and $i$-th vectors is not greater than  $\vc{N}_i=n^2/m_{\max}$ (see lines \ref{oldsel3}-\ref{oldsel5}). 
This procedure replaces the vector with the smallest temporal distance to its older neighbor by the most recent vector $\vc{p}^{(t+1)}_{c}$ (see line \ref{oldupdateM}). Periodically, the oldest vector is also removed to constrain the distance according to $\vc{N}$ (see line \ref{oldsel5}). 
The step-size adaptation of LM-MA-ES is the same as in MA-ES and CMA-ES, i.e., based on the cumulative step-size adaptation (CSA) rule of \citep{hansen1996adapting} (see line \ref{oldMAStepSizeUpdate}).

As it is mentioned in section 3 of the main paper, we found that instead of a rather complicated procedure described in this section, one can continuously update a set of $\vc{m}$ vectors that is distantly similar to computing momentum vectors in Stochastic Gradient Descent. The idea was born while attempting to implement Algorithm 1 for training deep neural networks. Since the performance of both algorithms is comparable, we decided to present the simpler one (given in the main paper) as our main method.

\setcounter{algorithm}{0}
\begin{algorithm}[tb!]
\caption{LM-MA-ES with the same procedure to store  direction vectors as in LM-CMA-ES}
\label{oldLMMA}
\begin{algorithmic}[1]
\STATE{\textbf{given} $n \in \mathbb{N}_+$, $m_{max} = 4 + \lfloor 3 \ln \, n  \rfloor $, $\lambda = 4 + \lfloor 3 \ln \, n  \rfloor $, $\mu =  \lfloor \lambda/2   \rfloor $, 
											$w_i = \frac{ \ln(\mu + \frac{1}{2}) - \ln\,i}{ \sum^{\mu}_{j=1}(\ln(\mu + \frac{1}{2})-\ln\,j)} \; \mstr{for} \; i=1, \ldots, \mu$,
											$\mu_w = \frac{1}{\sum^{\mu}_{i=1} w^2_i}$,
											$c_1 = \frac{1}{n}$, $c_{\sigma} = \frac{1}{m_{max}}$, $c_{c} = \frac{1}{n}$, 
													$d_{\sigma} = 0.5$,
											} \label{oldMAEScmaGiven}
\STATE{\textbf{initialize} $t \leftarrow 0, \vc{y}^{(t=0)} \in \R^{\dd}, m^{(t=0)} = 0,\sigma^{(t=0)} > 0, \vc{p}^{(t=0)}_{\sigma} = \ma{0}, \vc{p}^{(t=0)}_{c} = \ma{0}, \M \in \R^{m_{max} \times \dd }, \vc{ref} \in \mathbb{N}_+^{m_{max}}, \vc{time} \in \mathbb{N}_+^{m_{max}}, \vc{N} \in \mathbb{N}_+^{m_{max}-1} = n^2/m_{max}$}
\REPEAT
  \FOR{$i \leftarrow 1,\ldots,\lambda$} \label{oldMAGenerateBegin}
				\STATE{$\vc{z}^{(t)}_i \leftarrow {{\mathcal N}  \hspace{-0.13em}\left({\ma{0},\I\,}\right)}$}
				\STATE{$\vc{d}^{(t)}_i \leftarrow \vc{z}^{(t)}_i$}
				\FOR{$j \leftarrow m^{(t)},\ldots,1$} \label{oldLMloop1}
                		\STATE{$c^{'} \leftarrow c_1$}
                		\IF {$j < m^{(t)}$}
                			\STATE{$c^{'}_1 \leftarrow c_1 (\vc{time}_{\vc{ref}_{j+1}} - \vc{time}_{\vc{ref}_{j}})/\vc{N}_j$}
                        \ENDIF
						\STATE{$\vc{d}^{(t)}_i \leftarrow (1 - c^{'}_1) \vc{d}^{(t)}_i + c^{'}_1 \M_{\vc{ref}_j} \left(\sum_{k=1}^n \M_{\vc{ref}_j, k} \vc{d}^{(t)}_{i, k}\right)$}		 \label{oldLMloop2}
				\ENDFOR
			\STATE{ $\vc{f}^{(t)}_i \leftarrow f(\vc{y}^{(t)} + \sigma^{(t)} \vc{d}^{(t)}_i)$} 
  \ENDFOR
	\STATE{ $ \vc{y}^{(t+1)} \leftarrow  \vc{y}^{(t)} + \sigma^{(t)} \sum_{i=1}^{\mu} w_i \vc{d}^{(t)}_{i:\lambda} \;$} // the symbol $i:\lambda$ denotes $i$-th best sample on $f$ \label{oldMAComputeNewMean}
	\STATE{ $ \vc{p}^{(t+1)}_{\sigma} \leftarrow (1 - c_{\sigma}) \vc{p}^{(t)}_{\sigma} + \sqrt{\mu_w c_{\sigma}(2-c_{\sigma})} \sum_{i=1}^{\mu} w_i \vc{z}^{(t)}_{i:\lambda} $} \label{oldMASigmaPathUpdate}
	\STATE{ $ \vc{p}^{(t+1)}_{c} \leftarrow (1 - c_{c}) \vc{p}^{(t)}_{c} + \sqrt{\mu_w c_{c}(2-c_{c})} \sum_{i=1}^{\mu} w_i \vc{z}^{(t)}_{i:\lambda} $} \label{oldMASigmaPathUpdate2}
	\STATE{$m^{(t+1)} \leftarrow \mstr{min}(t+1,m_{max})$}
	\IF{$\left(t < m_{max}\right)$}
	\STATE{$ \vc{ref}_{m^{(t+1)}} \leftarrow t + 1$}
	\ELSE
	\STATE{$ i_{min} \leftarrow  argmin_i\left(\vc{time}_{\vc{ref}_{i+1}}-\vc{time}_{\vc{ref}_i} - \vc{N}_{i}\right),|1\leq i \leq (m^{(t+1)}-1)$}	\label{oldsel3}
		\IF{$ \left(\vc{time}_{\vc{ref}_{i_{min}+1}}-\vc{time}_{\vc{ref}_{i_{min}}} - \vc{N}_{i_{min}} \geq 0\right)$}	\label{oldsel4}
			\STATE{$ i_{min} \leftarrow 0$}	\label{oldsel5}
		\ENDIF{} 
		\STATE{remove $(i_{min}+1)$-th element of $\vc{ref}$ and insert it at now empty $m^{t+1}$-th  position} \label{oldsel6}
	\ENDIF
	\STATE{$\vc{time}_{\vc{ref}_{m^{(t+1)}}} \leftarrow t + 1$}
	\STATE{ $ \M_{\vc{ref}_{m^{(t+1)}}} \leftarrow \vc{p}^{(t+1)}_{c}$ }  \label{oldupdateM}
	\STATE{ $ \sigma^{(t+1)} \leftarrow \sigma^{(t)} \mstr{exp}
	          \left[ \frac{c_{\sigma}}{d_{\sigma}} \left(  \frac{\left\| \vc{p}^{(t+1)}_{\sigma} \right\|^2}{ n} - 1  \right) \right] $} \label{oldMAStepSizeUpdate}
  \STATE{ $ t \leftarrow t + 1$}
\UNTIL{ \textit{stopping criterion is met} }
\end{algorithmic}
\end{algorithm}

\section{Invariance}
\label{section:invariance}

LM-MA-ES, just like MA-ES, is invariant to translation, rotation and scaling of the objective function in search space, provided that the initial search distribution is transformed accordingly. However, in order to justify the use of separable test functions in the experimental evaluation, we validate this property empirically, by showing median runs (out of 5 runs) of the algorithm on separable and rotated problems. Figure~\ref{figure:rotated} shows convergence curves. The only systematic effect is due to initialization in the hypercube $[-5, 5]^n$, which is not rotated. The deviations are minimal, and no larger than the usual deviations due to randomized initialization and operation of the algorithm. This is in contrast to VD-CMA-ES \citep{akimoto2014comparison}, which is unable to solve rotated versions of some of the test problems.

\setcounter{figure}{0}
\begin{figure}[!ht]%
	\includegraphics[width=0.5\textwidth]{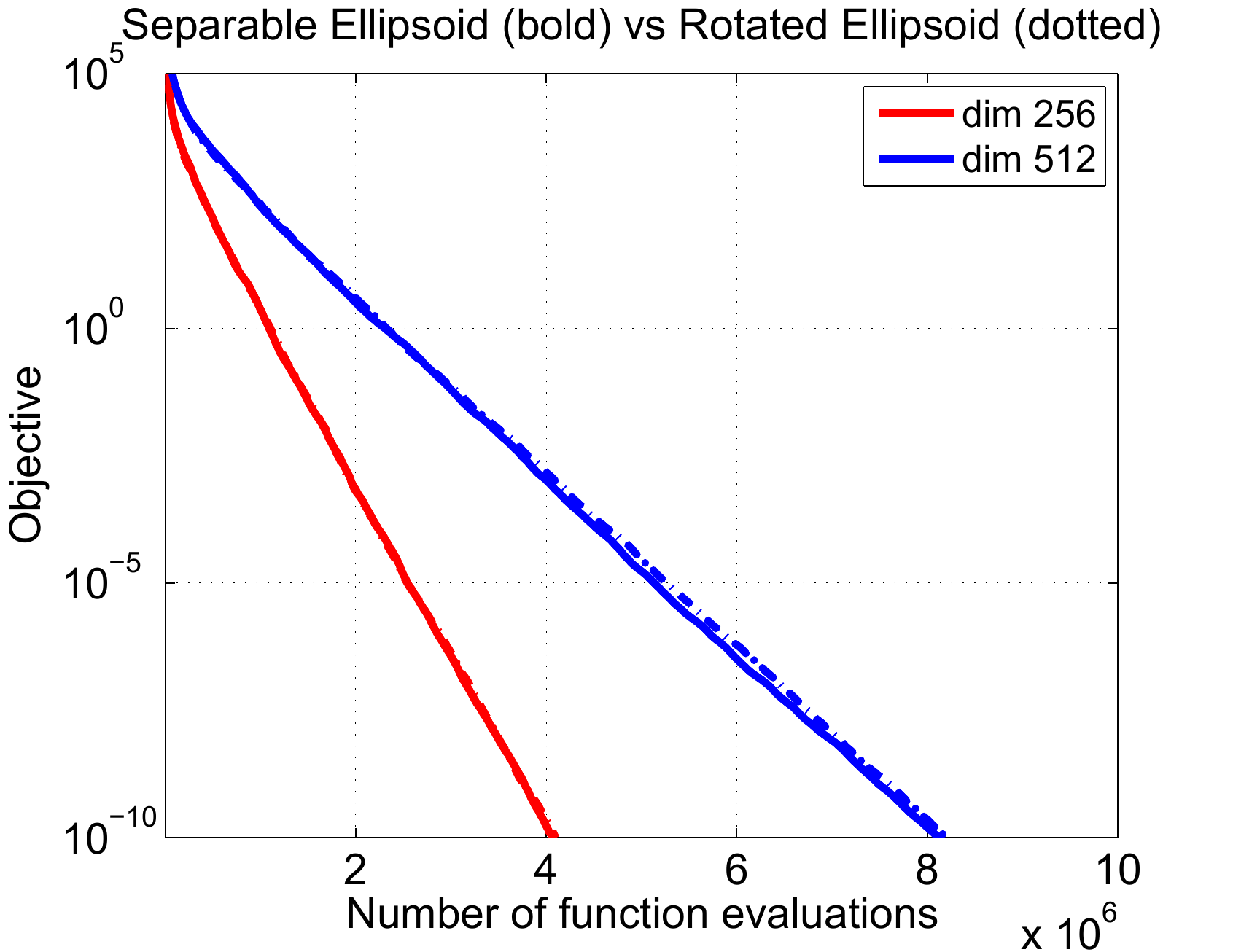}
	\includegraphics[width=0.5\textwidth]{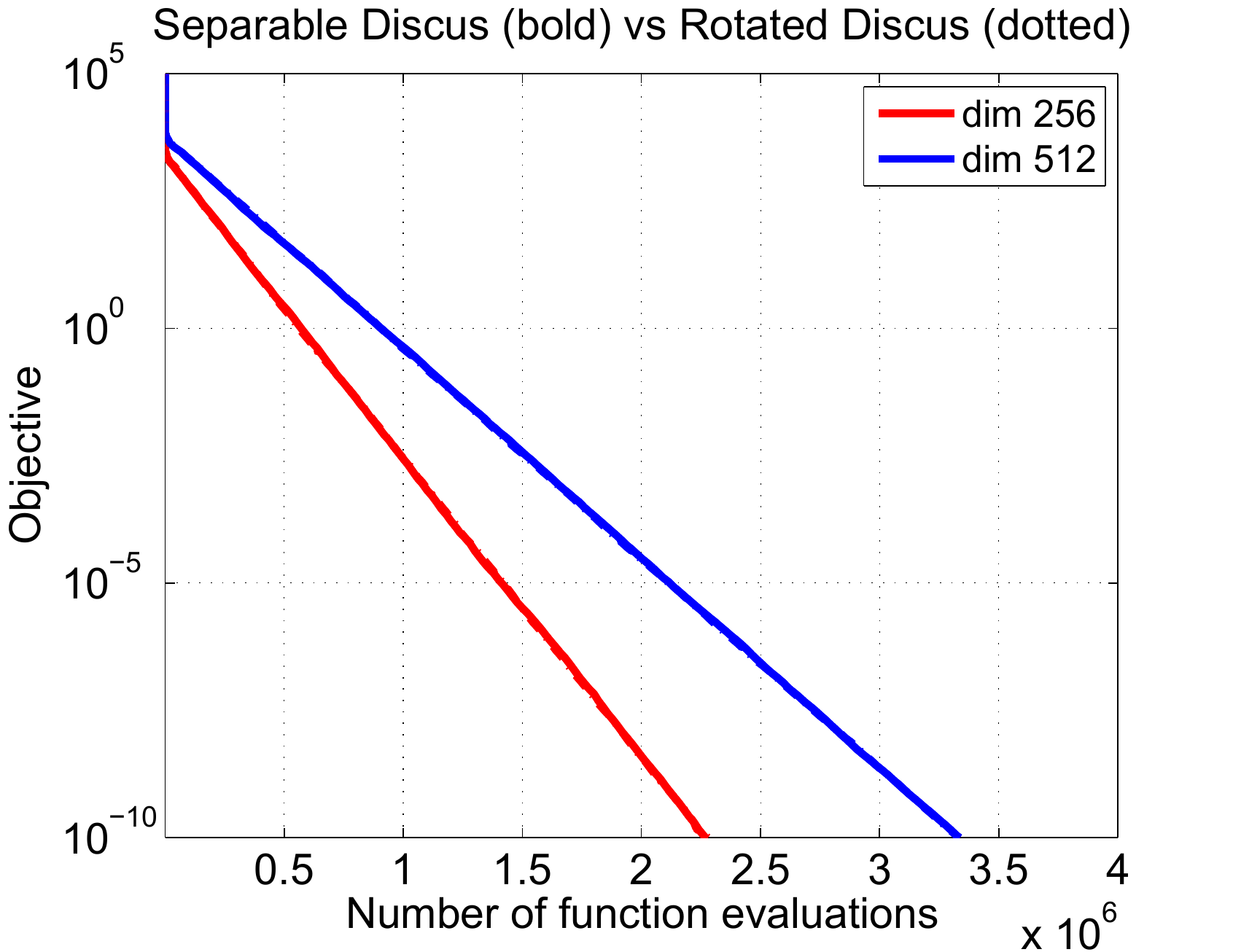}\\ 
  \includegraphics[width=0.5\textwidth]{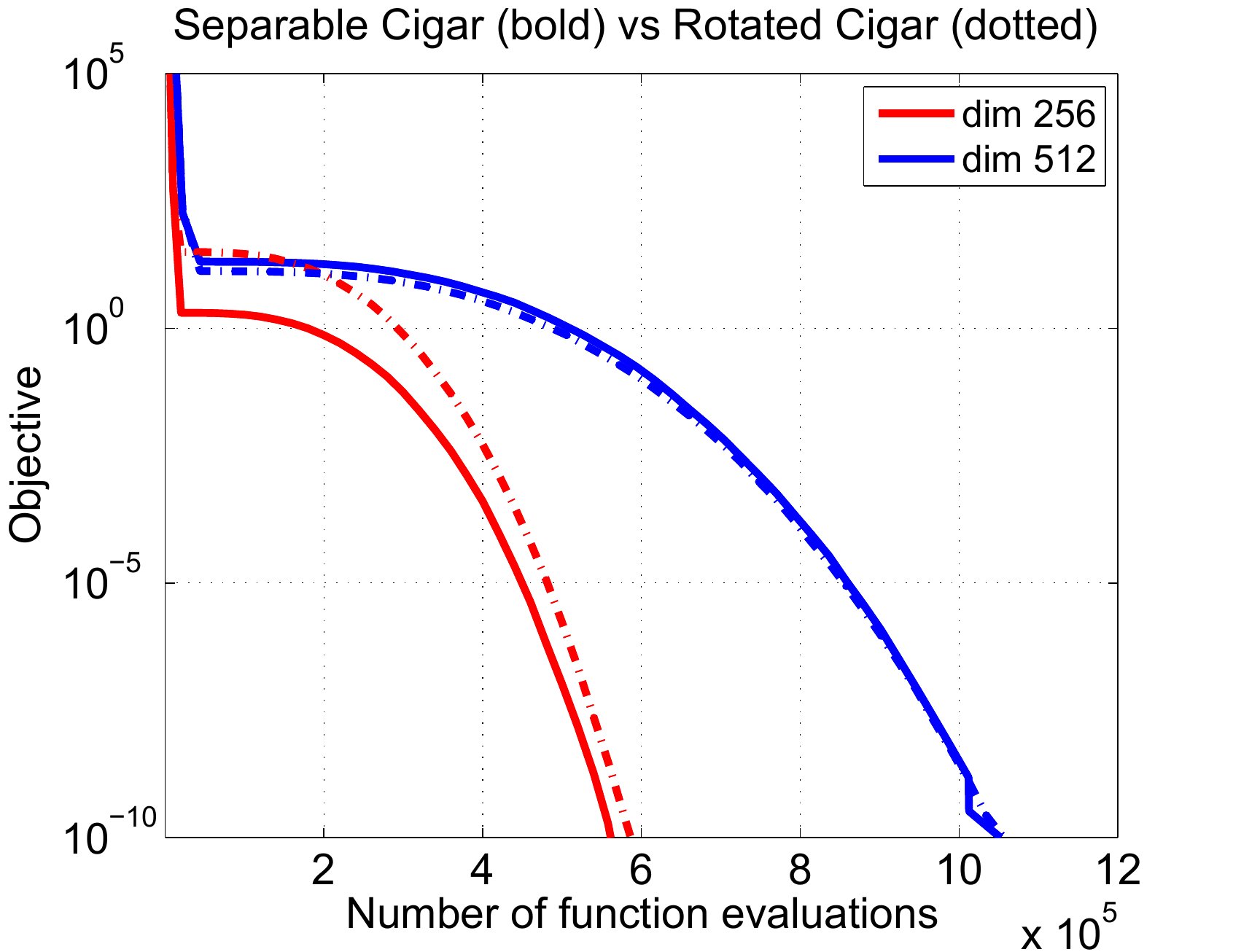}
	\includegraphics[width=0.5\textwidth]{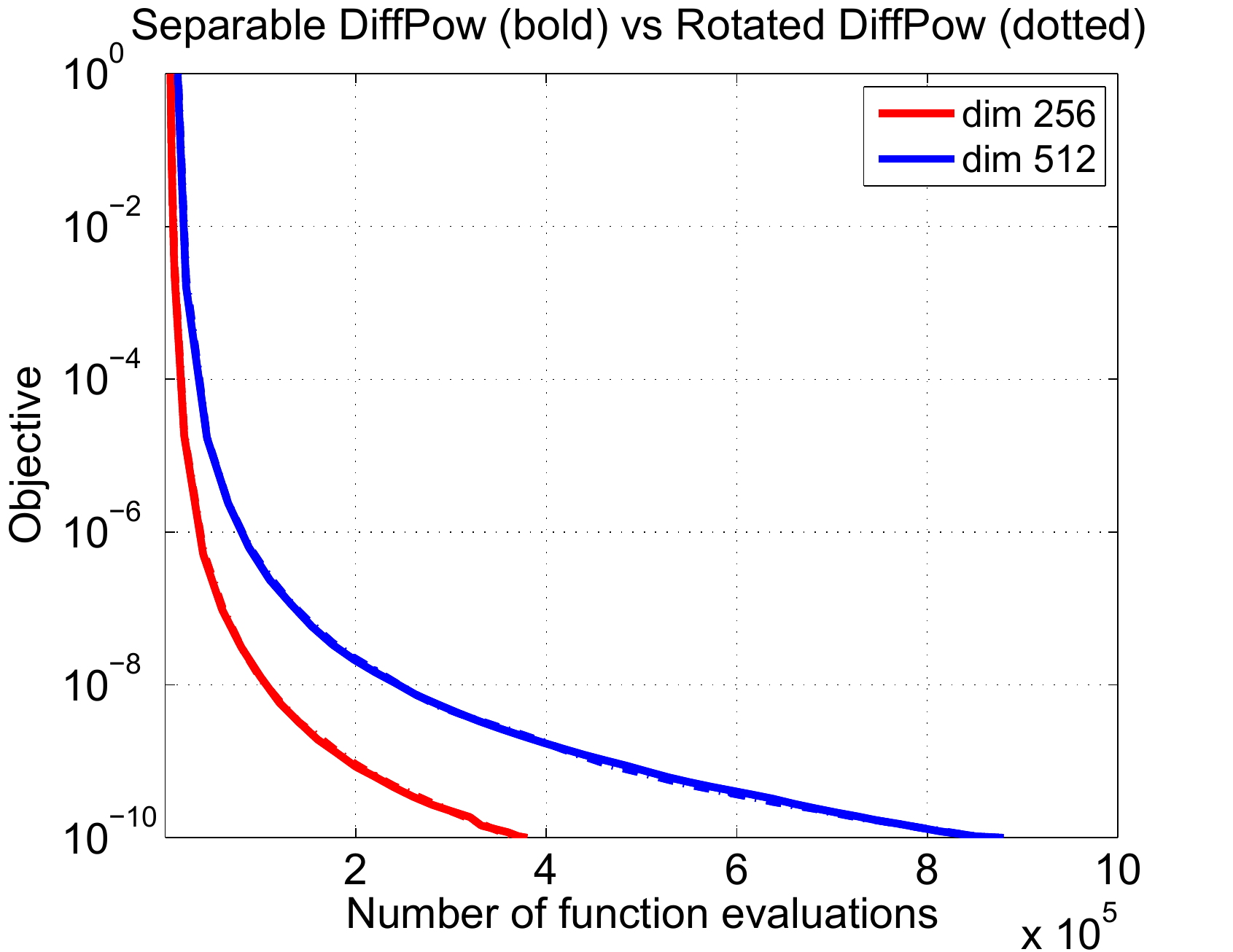} 
\caption{\label{figure:rotated} The trajectories show the median of 5   runs of LM-MA-ES on separable (bold lines) and rotated (dotted lines) versions of Ellipsoid, Discus, Cigar and Different  Powers functions (see Table 1 of the main paper) for problem dimensions 256 and 512. The rotated functions $f(\ma{R} \vc{x})$ are obtained from the original separable functions $f(\vc{x})$ by applying $\ma{R}$, an orthogonal $n \times n$ matrix with each column vector $\vc{q}_i$ being a uniformly distributed unit vector implementing an angle-preserving transformation. LM-MA-ES  numerically demonstrates invariance to rotations. The observed difference is due to the fact that the rotations effectively change the initial range of the search domain.}
\end{figure}

\end{document}